\documentclass[10pt,journal,cspaper,compsoc]{IEEEtran}
\usepackage{graphicx}
\usepackage{amsmath,amssymb}
\usepackage{color}
\usepackage{booktabs}
\usepackage{multirow}
\usepackage[caption=false]{subfig}
\usepackage{ragged2e}
\usepackage{multirow,array}
\usepackage{cite}
\usepackage[export]{adjustbox}
\usepackage{rotating}
\usepackage{multicol,multirow}
\usepackage{eqparbox}
\usepackage{overpic}
\newcommand{\todo}{}
\usepackage[colorlinks,bookmarksnumbered,bookmarksopen,citecolor=black]{hyperref}
\usepackage{silence}
 \def\zhou{}
 \def\zl{}
\def\cmm{}

\usepackage{xspace}

\def\ie{\textit{i.e.}}

\def\etal{\textit{et al. }}

\newcommand{\dimh}{\ensuremath{H}\xspace}
\newcommand{\dimw}{\ensuremath{W}\xspace}
\newcommand{\dimc}{\ensuremath{C}\xspace}

\newcommand{\lists}{\ensuremath{R}\xspace}
\newcommand{\feat}{\ensuremath{f}\xspace}

\newcommand{\counts}{\ensuremath{M}\xspace}
\newcommand{\Counts}{\mathcal{M}\xspace}

\newcommand{\data}{\ensuremath{X}\xspace}
\newcommand{\rnnoutput}{\ensuremath{o}\xspace}
\newcommand{\rnnoutstate}{\ensuremath{r}\xspace}
\newcommand{\rnnwt}{\ensuremath{W_{in}}\xspace}
\newcommand{\rnnws}{\ensuremath{W_s}\xspace}

\newcommand{\nvideos}{\ensuremath{N}\xspace}

\newcommand{\featSpace}{\ensuremath{\mathrm{\cal X}}\xspace}

\newcommand{\nframes}{\ensuremath{T}\xspace}

\newcommand{\samp}{\ensuremath{\mathbf{x}}\xspace}
\newcommand{\laby}{\ensuremath{\mathbf{Y}}\xspace}
\newcommand{\func}{\ensuremath{\mathbf{g}}\xspace}
\newcommand{\dist}{\ensuremath{\mathbf{d}}\xspace}

\newcommand{\shuffle}{\ensuremath{\mathcal{S}}\xspace}
\newcommand{\ranker}{\ensuremath{{R}}\xspace}

\newcommand{\RR}{I\!\!R}

\newcommand{\task}{\emph{VPRe-id}\xspace}
\newcommand{\ordered}{\emph{ordinal}~}
\newcommand{\inordered}{\emph{shuffled}~}

\graphicspath{{./image/}{./image/author/}}

\newcommand{\myPara}[1]{\vspace{0pt} \noindent {\bf #1.}}

\ifdefined \GramaCheck
  \newcommand{\CheckRmv}[1]{}
  \newcommand{\figref}[1]{Figure 1}%
  \newcommand{\tabref}[1]{Table 1}%
  \newcommand{\secref}[1]{Section 1}
  \renewcommand{\eqref}[1]{Equation 1}
\else
  \newcommand{\CheckRmv}[1]{#1}
  \newcommand{\figref}[1]{Fig.~\ref{#1}}%
  \newcommand{\tabref}[1]{Fig.~\ref{#1}}%
  \newcommand{\secref}[1]{Sec.~\ref{#1}}
  \renewcommand{\eqref}[1]{Eqn.~(\ref{#1})}
\fi

\begin{document}

\title{Ordered or Orderless: A Revisit for Video based Person Re-Identification}

\author{Le~Zhang,  Zenglin~Shi, Joey Tianyi Zhou, \\ 
  Ming-Ming Cheng, Yun~Liu, Jia-Wang Bian,  Zeng Zeng and 
  Chunhua Shen 
   
  \IEEEcompsocitemizethanks{\IEEEcompsocthanksitem 
    L. Zhang and Z. Zeng are with the Institute for Infocomm Research, the Agency for Science, 
    Technology and Research (A*STAR), Singapore.
    \IEEEcompsocthanksitem   J. T. Zhou is with the Institute of High Performance Computing, the Agency for Science, 
    Technology and Research (A*STAR), Singapore.
  \IEEEcompsocthanksitem Z. Shi is with the University of Amsterdam, Netherlands.
  \IEEEcompsocthanksitem J. W. Bian and C. Shen is with the School of Computer Science, The University of Adelaide, Australia.
  \IEEEcompsocthanksitem M. M. Cheng and Y Liu are with the TKLNDST, College of Computer Science, Nankai University, China.
  \IEEEcompsocthanksitem 
     The first two authors are the joint first author, and Joey Tianyi Zhou is the corresponding author (Email: joey.tianyi.zhou@gmail.com).}
  \thanks{Manuscript received April 19, 2005; revised August 26, 2015.}
}

\markboth{IEEE Transactions on Pattern Analysis and Machine Intelligence}%
{Zhang \MakeLowercase{\textit{et al.}}: Robust Regression via Deep Negative Correlation Learning}

\IEEEtitleabstractindextext{%
\begin{abstract}
\justifying
Is recurrent network really necessary for learning a good visual representation 
for video based person re-identification (\task)? 
In this paper, we first show that the common practice of employing recurrent 
neural networks (RNNs) to aggregate temporal-spatial~features may not be optimal.
Specifically, with a diagnostic analysis, \zl{we show that the recurrent structure 
may not be effective learn temporal dependencies than what we expected and implicitly yields an orderless representation}. 
Based on this observation, 
we then present a simple yet surprisingly powerful approach for \task,
where we treat \task as an efficient \zl{orderless} ensemble of image based person 
re-identification problem. 
More specifically, we divide videos into individual images and 
re-identify person with ensemble of image based rankers. 
Under the i.i.d. assumption, 
we provide an error bound that sheds light upon how could we improve \task.
%
Our work also presents a promising way to bridge the gap between video and image
based person re-identification. 
Comprehensive experimental evaluations demonstrate that the proposed solution 
achieves state-of-the-art performances on multiple widely used datasets 
(iLIDS-VID, PRID 2011, and MARS).

\end{abstract}

}

\maketitle
 
\IEEEdisplaynontitleabstractindextext

\IEEEpeerreviewmaketitle

\section{Introduction}\label{sec:introduction}

\IEEEPARstart{P}{erson} re-identification (Re-id) addresses the problem 
of re-association persons across disjoint camera views. 
In this paper, we consider more practical scenarios of video based person 
re-identification (\task), 
in which a video of a person, as seen in one camera, 
must be matched against a gallery of videos captured by a different 
non-overlapping camera. 
\task is an active research topic in computer vision due to its wide-ranging
applications in problems, 
including visual surveillance and forensics. 
Since the pioneering work \cite{gheissari2006person}, 
several visual features \cite{farenzena2010person,kviatkovsky2013color}, 
and learning methods \cite{hirzer2011person,hirzer2012relaxed,yan2016person} 
have consistently improved the matching performance, 
leading the research community to address more challenging scenarios in 
complex datasets \cite{hirzer2011person,wang2014person,zheng2016mars}.
However, significant hurdles due to variations in appearance, viewpoint,
illumination, and occlusion come in the way of solving the problem.

\begin{figure}[!tb]
  \centering
  \includegraphics[width=0.4\textwidth]{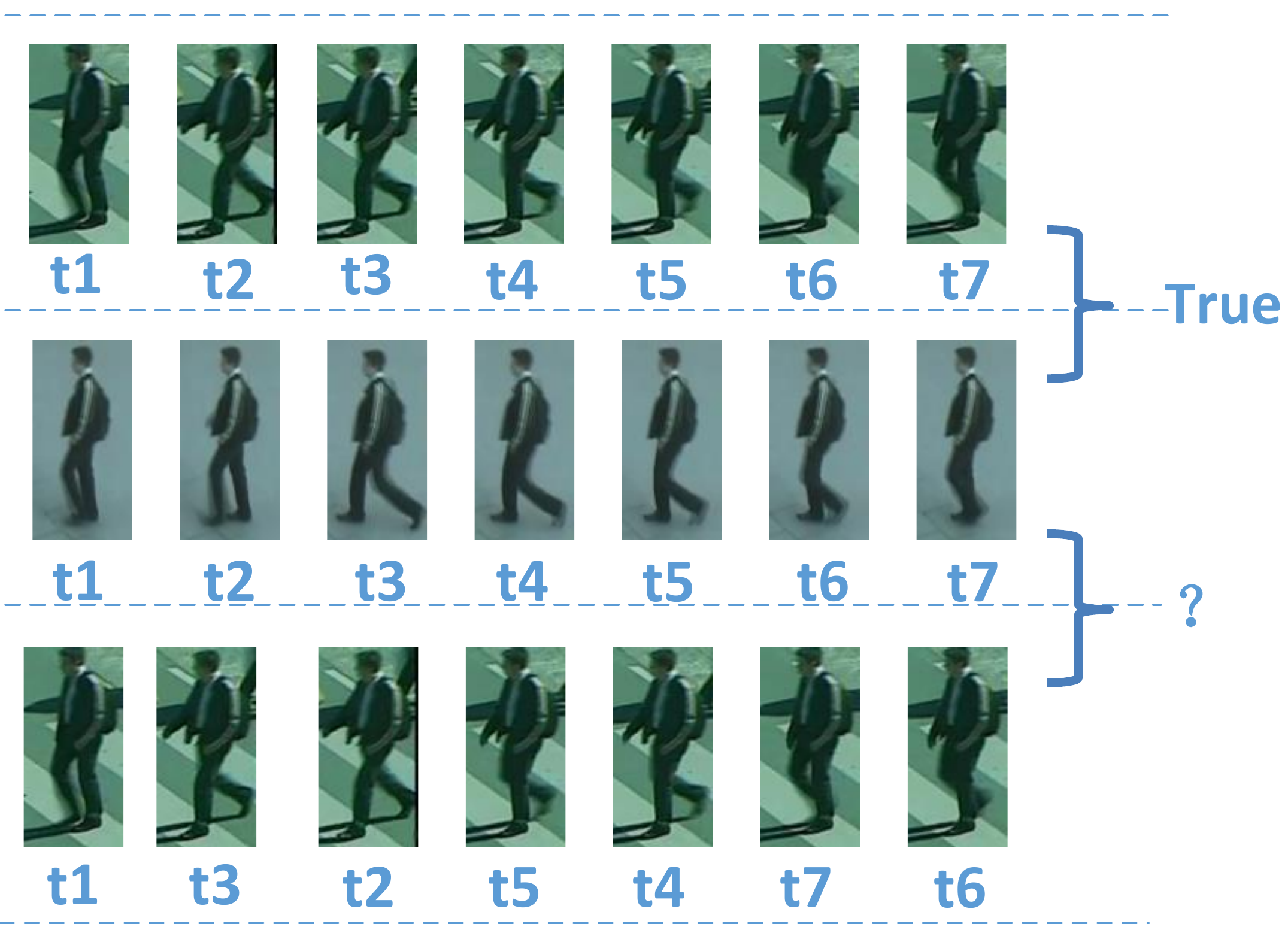}
  \caption{\textbf{Motivation.} 
    First two rows: existing methods adopts RNNs to model temporal 
    dependencies for~\task. 
    Last two rows: human beings can easily performs this on image sequences 
    with a random order.
  }\label{fig:reidquestion}
\end{figure}

Recently, deep convolutional neural networks (ConvNets) stand at the 
forefront of several vision tasks, 
including image classification \cite{krizhevsky2012imagenet,GoogleNet,VGG,Resnet},
segmentation \cite{long2015fully}, 
pose estimation~\cite{Tompson:pose:cvpr15}, 
face recognition \cite{Schroff:cvpr:2015}, 
crowd counting~\cite{shi2018crowd}, 
and image based person re-identification~\cite{li2014deepreid}, 
just to mention a few. 
Deep learning for \task, however, is also witnessing such vast popularity. 
Typically, the prior methods
\cite{mclaughlin2016recurrent,yan2016person,wu2016deep,chen2017deep,xu2017jointly,zhousee,chen2018video} 
process videos by using RNNs to temporally aggregate spatial information 
extracted from ConvNets.

However, unlike other sequential modelling tasks~\cite{caba2015activitynet}, 
it is little-known about the benefits of using the RNN for 
modelling temporally extracted spatial information for~\task.  
Although theoretically fascinating, 
we show a typical recurrent structure may \zl{ be less effective to capture temporal dependencies 
as they assumed}. 
To simplify our analysis, we find that the pioneering work
\cite{mclaughlin2016recurrent} of using RNN and its follow-ups
for~\task leads to an orderless representation. 
This motivates us to re-ponder over the task of~\task as illustrated in
\figref{fig:reidquestion}.  
According to the separable visual pathway hypothesis~\cite{goodale1992separate}, 
the human visual cortex contains the ventral stream and the dorsal stream, 
which recognize objects and how they move, respectively. 
Existing methods ~\cite{mclaughlin2016recurrent} employing RNNs overemphasize the 
temporal dependencies related to person's motion. 
Unfortunately, these behavioral biometrics suffer from large 
\zl{intra/inter-class variations}, 
making existing RNNs difficult to generalize. 
In this work we postulate that the appearance information from a single still 
image plays a more important role in re-associating persons from different cameras.
Considering an example in \figref{fig:reidquestion}, 
human beings can easily classify whether the images in two cameras come from 
the same identity, 
even the image frames in the second camera has been manually shuffled to 
remove temporal dependencies. 

Declaring that orderless encoding's benefits in video analytic is 
non-controversial, and indeed we are certainly not the first to 
inject orderless encoding into video based tasks. 
Even in other tasks such as human action recognition where the temporal 
dependency between each video frame is considered to be vital, orderless
representations \cite{wang2016temporal,girdhar2017actionvlad,zhu2016key} 
can still demonstrate their superiority over recurrent 
structures~\cite{donahue2015long}. 
While we take inspiration from these works and the recent success of 
image based person re-identification, 
we are the first to dive deep into the effect of modelling 
temporal information for \task. 
More specifically, we present a simple yet surprisingly powerful approach 
by dividing videos into individual images and perform \task 
with
an 
\zl{orderless} ensemble of shared image based rankers. 
We show that, under the i.i.d.\   assumption,  
the ensemble of rankers are consistent and the error bound of \task 
decreases exponentially with the number of image frames, 
when each base ranker is better than a random guess. 
Our work opens a path towards putting more emphasis on appearance information 
for \task. 
It not only achieves state-of-the-art performance on multiple benchmarks 
(iLIDS-VID, PRID 2011, and MARS) but also 
bridges the gap between video and image based person re-identification.

\section{Related Work}
%
%

\subsection{Recurrent Structures for \task}

Recurrent structures, e.g., RNNS, LSTMs, GRUs, etc., have been widely used to
temporally aggregate spatial information extracted from each video frame.
Yan \etal \cite{yan2016person} adopts \emph{ Long-Short Term Memory networks} 
(LSTM) to sequentially fuse hand-crafted features. 
In the same way, recurrent networks~\cite{mclaughlin2016recurrent} are used 
to aggregate discriminative features extracted from other ConvNets on 
individual video frames. 
Wu \etal \cite{wu2016deep} consider convolutional activations at low levels, 
which are embedded into \emph{Gated Recurrent Units} (GRU) to 
capture temporal patterns. 
\emph{Bidirectional RNNs} are introduced in~\cite{zhang2017learning} 
to integrate convolutional features. 
\emph{Deep Feature Guided Pooling} (DFGP) is proposed in~\cite{li2017video} 
to jointly 
%
%
use
both deep features and LOMO hand-crafted features. 
A similar ``CNN-RNN" recipe is trained in~\cite{chen2017deep}  
on full body and part-body image sequences respectively to 
learn complementary representations from holistic and local perspectives.
\emph{Spatial and Temporal Attention Pooling Network} (ASTPN)~\cite{xu2017jointly}
extends the standard CNN-RNNs by decomposing pooling into spatial-pooling 
on feature maps from CNN and an attentive temporal-pooling on the output of RNN.
Spatial recurrent models and temporal attention models are proposed in
\cite{zhousee} to both select discriminative frames and 
exploit the contextual information when measuring the similarity. 
A recurrent based competitive snippet-similarity aggregation and co-attentive 
snippet embedding is introduced in~\cite{chen2018video}. ~\zhou{\cite{su2018spatial} extracts spatial-temporal features using a novel recurrent-based spatial temporal synergic residual network.}

\subsection{Image based Person Re-id}

Compared with~\task, image based person re-identification has been more 
extensively studied. 
A filter pairing neural network is designed in~\cite{li2014deepreid} 
to jointly handle misalignment and geometric transformations. 
A cross-input difference CNN is proposed in~\cite{ahmed2015improved} 
to capture local relationships between the two input images based %
on mid-level features from each input image. 
Xiao~\etal introduces a domain guided drop-out technique to mitigate the domain 
gaps from various datasets \cite{xiao2016learning}. 
Some other works aim at designing more robust loss function such as the 
triplet loss \cite{ding2015deep,cheng2016person,zhao2017deeply} 
to enforce the correct order of relative distances among image triplets 
or quadruplet loss \cite{chen2017beyond} to combine the the advantages of contrastive loss and triplet loss. 
The similarities between different gallery images are reported to be beneficial 
for re-identification in \cite{shen2018deep,chen2018group}. 
\zhou{Song~\etal~\cite{song2018mask} propose a novel region-level triplet loss to pull the features from the full image and
body region close, whereas pushing the features from backgrounds away.}
\subsection{Bridging the Research Gap} 
Despite the abundant work published 
using RNNs for \task, almost none of them has  demonstrated that  the improved accuracy of \task using  RNNs  
can indeed be traced back to that the recurrent structure encodes temporal information that is missing in image based person re-id.  
 Instead, the community have largely taken it for granted that recurrent structures are able to successfully capture
the temporal dependency in \task.  
To investigate the effect of using recurrent structures for \task, 
we make the first attempt to investigate the effect of leveraging temporal
information by 
using
recurrent structures for \task. 
\zl{Please note that other non-recurrent method for \task include 
\cite{zhang2018multi}, 
which solves~\task by 
using
reinforcement learning. 
However, they do not study the effect of the frame order in a systematic manner. 
The other example 
is
\cite{boin2019recurrent}, 
in which the authors derive an approximation of the recurrent connections 
and achieve faster and competitive results with a simpler feed-forward architecture.  Other methods~\cite{zhousee} try to use recurrent structures to model context information \zhou{whose effectiveness has been verified recently~\cite{si2018dual}}. However, they still lack a systematical study to demonstrate the capability of RNN's context modelling with respect to frame order.}
Our diagnostic analysis, as demonstrated in \secref{sec:diag}, 
shows the deficiency of the commonly used ``CNN-RNN" pipeline: 
the recurrent structure \zl{is less effective in modelling the temporal information than what we expected}.
We then speculate that orderless approaches may be more pertinent for the task 
and introduce a simple yet powerful solution to bridge these two research domains. 
We formulate \task as an \zl{orderless} ensemble of rankers and employ image based person 
re-identification methods as base rankers on each individual frame. 
Theoretical analysis is provided to further guarantee the performance of the 
proposed methods. 
Our proposed approach differs significantly from 
previous work
that %
is based on recurrent structures, 
as 
our method does not rely on
the temporal information in the video.  


\section{Notation and Problem Definition}

Denoted by $\{\data^{p_i},\data^{g_i}\}_{i=1}^\nvideos$, 
a collection of pedestrian videos, 
where $\data^{p_i}$ (probe video) and $\data^{g_i}$ (gallery video) 
stand for the $i^{th}$ pedestrian video captured by two disjoint cameras 
$p$ and $g$, and $\nvideos$ is the number of videos%
\footnote{For simplicity here we consider only two camera views are available 
and the extension to more views is straightforward.}. 
Each video is defined as a set of consecutive frames, \ie,  
$\data^{p_i}=[\samp_{1}^{p_i},\samp_{2}^{p_i},\cdots,\samp_{\nframes^{p_i}}^{p_i} ]$,
where $\nframes^{p_i}$ is the number of frames for the $i^{th}$ video captured 
by camera $p$.  
We will denote a generic data point as $\samp$ by ignoring the subscript 
and superscript for convenience. 
In the same way, a generic video input sequence is denoted by 
$\data=[\samp_1,\samp_2,\cdots,\samp_\nframes]$. 
We consider that the video frames are present in a 
$\dimh \times \dimw \times \dimc$ dimensional space: 
$\samp \in \featSpace \subseteq \RR^{\dimh \times \dimw \times \dimc}$, 
where $\dimh$, $\dimw$, and $\dimc$ denote the height, width, 
and number of channels for an input $\samp$, respectively.

We now present the objective of \task and several definitions that will be 
frequently used later. 

\myPara{\task Objective} 
\emph{~Our objective is to classify whether the videos $\data^p$ and $\data^g$ 
are coming from the same identify. 
This is usually achieved by learning a mapping function $\func$ such that}: 
\begin{equation} \label{problem_define}
\begin{aligned}
  &\dist(\func(\data^{p_i}), \func(\data^{g_i}))
  < \dist(\func(\data^{p_i}), \func(\data^{g_j})),\\
  &\forall j \neq i~{\rm and}~ i,~j=1,2 \cdots \nvideos,
\end{aligned}
\end{equation}
\emph{where $\dist$ denotes a distance measurement such as the commonly 
used Euclidean Distance}.

\myPara{Definition 1} \emph{The following network is called a typical RNN:}
\begin{equation} \label{RNN}
\begin{aligned}
\rnnoutput_{(t)}=\rnnwt \feat(\samp_{t})+\rnnws\rnnoutstate_{(t-1)},\rnnoutstate_{(t)}=Tanh(\rnnoutput_{(t)}),
\end{aligned}
\end{equation}
\emph{where the output $\rnnoutput$ at each time-step $t$ is a linear combination
of the input $\feat(\samp_{t})$ and RNN's state $\rnnoutstate_{(t-1)}$ at 
previous time-step. 
$\feat$ stands for a generic feature extractor. $\samp_{t}$ stands for the 
$t^{th}$ frame for a generic video sequence.}

\myPara{Definition 2} 
\emph{An~\ordered video sequence is defined as $\data^{p_i}=[\samp_{1}^{p_i},
\samp_{2}^{p_i}, \cdots, \samp_{\nframes^{p_i}}^{p_i}]$. 
A~\inordered video sequence is defined as 
$\data^{p_i}=[\samp_{\shuffle(1)}^{p_i},\samp_{\shuffle(2)}^{p_i},\cdots,
\samp_{\shuffle(\nframes^{p_i})}^{p_i}]$, 
where $\shuffle$ is a random permutation operator that randomly permutates 
(or shuffles) the order of frames in the video sequence.}

\newcommand{\addImg}[2]{\subfloat[#1]{\includegraphics[height=0.19\textwidth]{#2}}}


\begin{figure}
  \centering
  \vspace{-.1in}
  \includegraphics[width=0.3\textwidth]{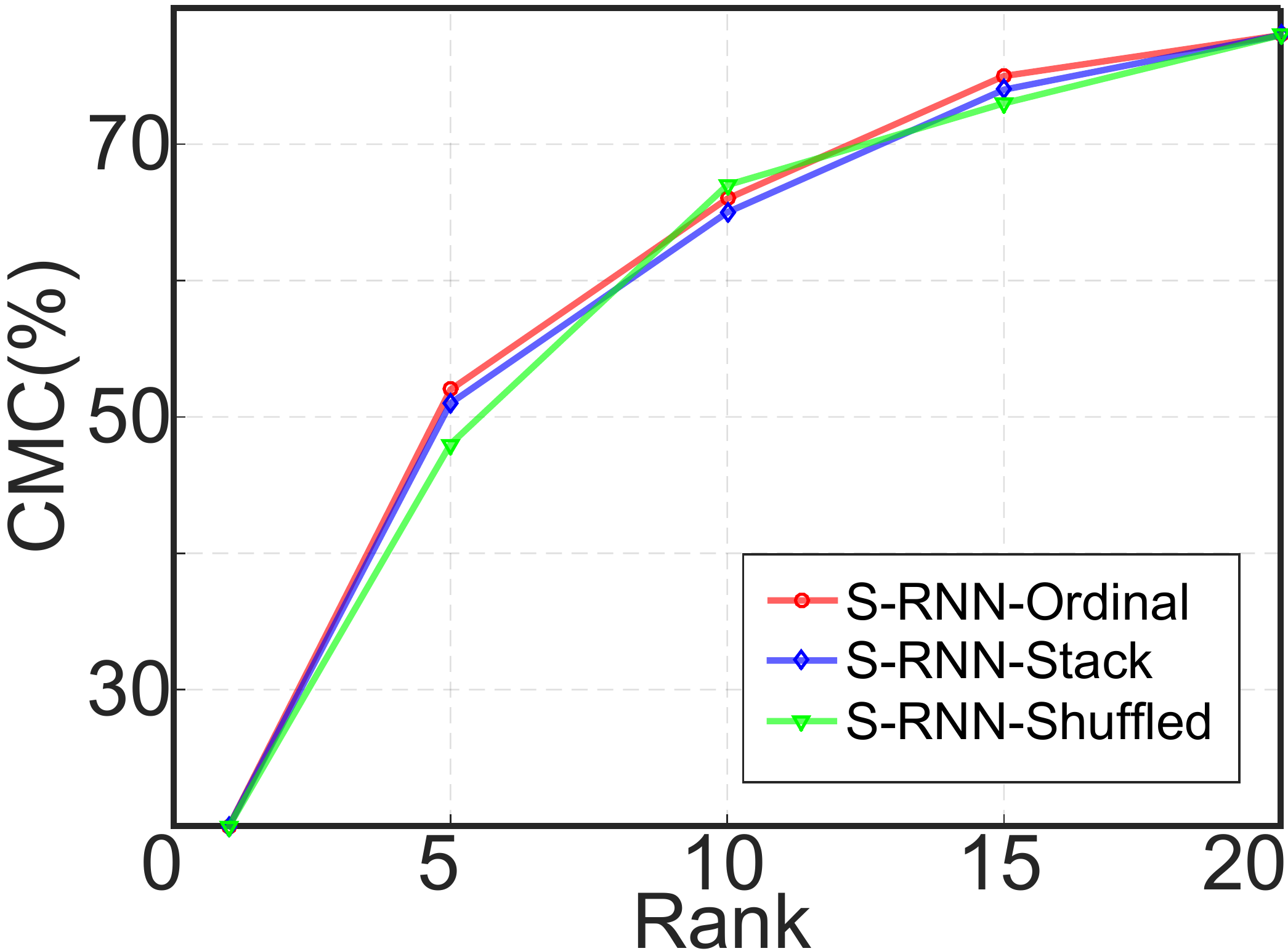}
  \caption{\zl{\textbf{Diagnostic analysis of a
    typical ``CNN-RNN" framework} in \cite{mclaughlin2016recurrent}.
  we consider a single RNN without temporal average pooling. 
	``S-CNN-RNN-Ordinal" and  ``S-CNN-RNN-Shuffled" stands for a single RNN 
	with ordinal, shuffled videos, respectively. 
	``S-CNN-RNN-Stack" means RNN with a stack of the same frames. 
  }}\label{fig:diag}
\end{figure}
\section{Diagnostic Analysis of ``CNN-RNN" }
\label{sec:diag}

As a proof of concept, 
firstly a diagnostic analysis of recurrent network is provided for \task. 
Our analysis is conducted based on the method presented in
\cite{mclaughlin2016recurrent} that has been widely adopted in many 
of its follow-ups works
\cite{zhang2017learning,wu2016deep,yan2016person,chen2017deep,zhousee,xu2017jointly}.
We will 
omit
the superscript of $\data$ to represent a video sequence from 
a generic camera view for convenience. 

\myPara{Observation 1} 
\emph{A typical CNN-RNN method, as done in ~\cite{mclaughlin2016recurrent}, 
\zl{is not effective in capturing temporal dependencies as it expects}, 
and implicitly results in an orderless representation.}

We illustrate this by \textbf{evaluating (without retraining)} 
the ``CNN-RNN" architecture on both \ordered and \inordered video sequences. 
The results are shown in \tabref{tab:lstm-gru}. 
It is surprising to see that commonly used ``CNN-RNN" architecture performs better 
on \inordered video sequences. 
This suggests that the sequential dependency assumption in video frames 
might be invalid for \task, 
so that the RNN learns to forget previous state. 
In this case, there are much more diversified input images in a~\inordered scenario.
For example, considering a video sequence with $\nframes$ frames and 
an RNN receiving $t$ frames in total as input, 
there exist $C_\nframes^t$ input proposals in \inordered scenario,
whereas in \ordered cases, one may only have $\nframes-t+1$ input proposals.

\newcommand{\colsN}[2]{\multicolumn{#1}{c}{#2}}
\newcommand{\rowsN}[1]{\multirow{4}{*}{#1}}

\begin{figure}[t]
  \scriptsize
  \centering
  \setlength{\tabcolsep}{0.9mm}
  \begin{tabular}{@{}ccccccccc@{}} \toprule
    &\colsN{2}{Train} & \colsN{2}{Test} & \colsN{4}{CMC Rank} \\
      \cmidrule(lr){2-3} \cmidrule(lr){4-5} \cmidrule(lr){6-9}
    & Ordinal & Shuffled & Ordinal & Shuffled & 1 & 5 & 10 & 20 \\
	\toprule  
    \rowsN{\cite{mclaughlin2016recurrent}}  
    & $\checkmark$ & &$\checkmark$ & & 58$\%$ & 84$\%$ & 91$\%$ & 96$\%$ \\
    & $\checkmark$ & & &$\checkmark$ & \textbf{59$\%$} & \textbf{88$\%$} & \textbf{94$\%$} & \textbf{97$\%$} \\
    & &$\checkmark$ & $\checkmark$ & & 58$\%$ & 87$\%$ & 92$\%$ & \textbf{97$\%$} \\
    & &$\checkmark$ & & $\checkmark$ & 58$\%$ & 87$\%$ & 92$\%$ & \textbf{97$\%$} \\    
    \toprule
    \rowsN{RNN}
    & $\checkmark$ & & $\checkmark$ & & 46$\%$ & 74$\%$ & 85$\%$ &94$\%$ \\
    & $\checkmark$ & & & $\checkmark$ & 46$\%$ & 74$\%$ & 86$\%$ &94$\%$ \\
    & & $\checkmark$ & $\checkmark$ & & 54$\%$  & \textbf{78$\%$} & \textbf{88$\%$} & \textbf{95$\%$} \\
    & & $\checkmark$ & & $\checkmark$ & \textbf{56$\%$}  & \textbf{78$\%$} & 87$\%$ & 94$\%$ \\
    \hline
    \rowsN{LSTM}
    & $\checkmark$ & & $\checkmark$ & & 41$\%$ & 68$\%$ & 82$\%$ & 91$\%$ \\
    & $\checkmark$ & & & $\checkmark$ & 44$\%$ & 72$\%$ & 84$\%$ & 91$\%$ \\
    & & $\checkmark$ & $\checkmark$ & & 49$\%$ & 73$\%$ & 85$\%$ & 92$\%$ \\
    & & $\checkmark$ & & $\checkmark$ & \textbf{52$\%$} & \textbf{76$\%$} & \textbf{86$\%$} & \textbf{93$\%$} \\
    \hline
    \rowsN{GRU} 
    & $\checkmark$ & & $\checkmark$ & & 44$\%$ & 74$\%$ & 83$\%$ & 93$\%$ \\
    & $\checkmark$ & & & $\checkmark$ & 46$\%$ & \textbf{76$\%$} & 85$\%$ & 92$\%$ \\
    & & $\checkmark$ & $\checkmark$ & & 51$\%$ & 74$\%$ & 86$\%$ & 93$\%$ \\
    & & $\checkmark$ & & $\checkmark$ & \textbf{54$\%$} & 74$\%$ & \textbf{87$\%$} & \textbf{94$\%$} \\
    \bottomrule
  \end{tabular}
  \caption{\zl{\textbf{Detailed results of different ``CNN-RNN" recipes}
    under different training and testing settings. \zhou{For the results of ~\cite{mclaughlin2016recurrent}, we follow the original protocol and use the pre-computed optical flow using the original ordinal video sequences.} For the remaining methods, the optical flow input is disabled.}}\label{tab:lstm-gru}
\end{figure}

We further show that a single RNN in~\cite{mclaughlin2016recurrent} 
usually leads to a negligible performance difference in various input scenarios 
such as \ordered, \inordered and stack of the same frame. 
More specifically, we remove the temporal pooling and take the last output 
of the RNN in \cite{mclaughlin2016recurrent} to further understand the merits 
of RNNs for \task. 
In the first two scenarios, we feed an \ordered and \inordered video frames 
($128$ video frames, as done in \cite{mclaughlin2016recurrent}) into RNN. 
In the third scenario, we \todo{repeatedly} stack one randomly selected video 
frame $128$ times to generate a ``fake" video snippet. 
All the experiments are repeated ten times and results are illustrated in 
\figref{fig:diag}. 
It can be seen that RNN leads to almost the same results, 
which implies that all previous inputs for RNN, 
whether it is in \ordered or \inordered manner, 
or even they are exactly the same, is less important. 
One may argue that forgetting its previous state may be the algorithmic 
deficiency of RNN. 
We empirically find that other advanced learning methods, such as LSTM, 
also lead to similar results. 
For more detailed results, please refer to \tabref{tab:lstm-gru}.

\myPara{Observation 2} 
\emph{The same CNN-RNN methods, when trained on randomly shuffled video sequences,
perform comparably good as those trained on ordinal video sequences.}

In order to further understand the merit of ``CNN-RNN" and clarify our motivation, 
we re-train the ``CNN-RNN" \cite{mclaughlin2016recurrent} system, 
but randomly shuffle the input videos in each mini-batch. 
We use the publicly  available implementation of \cite{mclaughlin2016recurrent}. 
The results are summarized in \tabref{tab:lstm-gru}. 
We also include the results of the above analysis for more detailed comparisons. 
In the first setting, we include the results of ~\cite{mclaughlin2016recurrent}. 
In this setting, the ``CNN-RNN" is trained on the original ordinal video sequences
and also evaluated on the ordinal ones. 
The second setting of ``Train-ordinal, Test-shuffled"  summarizes our previous
investigations. 
That is, we train ``CNN-RNN" on the original ordinal video sequences 
and evaluate it on randomly shuffled ones. 
In the last two rows, we randomly shuffle the video sequences in each 
epoch during the training stage. 
After that, we test the network on both randomly shuffled videos 
and the original ordinal sequences. 
Our results show that RNN fails to capture temporal
dependencies as it expects.

\zl{One may argue that the optical flow used in \cite{mclaughlin2016recurrent} 
may play an essential role in these settings. 
To investigate this, we conduct another set of experiments, 
in which the optical flow input is entirely disabled. 
Furthermore, more advanced recurrent structures such as GRU and LSTM are 
investigated as well. 
What we observe are: 
i) The optical flow is beneficial for this task, 
especially when the system is trained in ordinal sequences. 
However, this advantage becomes less apparent when the system is trained on 
the shuffled sequences. 
ii) the frame order, which is widely believed to encode temporal information 
for \task, 
is not beneficial for all three recurrent structures studied here.
For more observations and detailed analysis on the statistical significance test, 
please refer to the supplementary files.}


\section{Proposed Solution}
\subsection{Ensemble Ranker}
\zl{ It has been widely believed that the temporal information encoded in the frame order is vital for ~\task and hence, the community have
largely taken it for granted that using recurrent structures are good practices. For the first time,  we dive deep into the
effect of modelling temporal information for VPRe-id.
Our analysis shows that the commonly used recurrent structure could be less effective as what we expected in capturing sequential dependencies for~\task.
Indeed, it \emph{essentially} learns an orderless representation, 
which we believe could be more pertinent for~\task, 
\zl{by using multiple feature representations from the temporal pooling 
on RNN's output at different time-steps.} 
Moreover, in each time step, 
the output of RNN is dominated by the current input features. }

\zl{Motivated by this, we believe that it is beneficial to \emph{explicitly} 
solve \task in an orderless manner. This further motivates us to solve ~\task from a different angle. We propose a simple yet powerful solution by regarding \task as a task of orderless ensemble ranking where each base ranker is embodied 
with a person re-identifier with a single image frame. }
Multiple ranking results can be aggregated by the \emph{Kemeny–Young method}
\cite{levin1995introduction}. 
From ensemble learning point of view~\cite{ren2016ensemble}, 
strength and diversities are both essential, 
and in our case they are guaranteed by the accurateness of image based person 
re-identification approaches and differences of image frames within a video, respectively. 
Our method provides much more diversified \textbf{image} pairs than 
\textbf{video} pairs in the conventional method. 
All those image pairs contribute to the final re-id accuracy by providing 
a more detailed ranking under the KemenyYoung method, 
and they are all well-trained in our setting. 
Through two examples in \secref{sec:baseranker}, 
we show that the proposed solution can be easily integrated 
with any image based person re-identification approaches.

\myPara{Rationale} 
More formally, for each input video  
$\data^{p_i}=[\samp_{1}^{p_i}, \samp_{2}^{p_i}, \cdots, \samp_{\nframes^{p_i}}^{p_i}]$ 
in the probe set, 
our solution returns an ensemble of ranking list, composed of $\nframes^{p_i}$ 
base lists $\lists= \{\lists_t\}_{t=1}^{\nframes^{p_i}}$, 
where the ranking $\lists_t$, are generated by a base ranker $\lists$ 
whose final objective is:
\begin{equation}
\begin{aligned}
  &\dist(\ranker(\samp_{t}^{p_i}), \ranker(\samp_{t'}^{g_i}))
  < \dist(\ranker(\samp_{t}^{p_i}), \ranker(\samp_{t''}^{g_j}))),\\
  &\forall j \neq i, \forall  t \in \{1,2,\cdots,\nframes^{p_i}\},\\
  &\forall t' \in \{1,2,\cdots,\nframes^{g_i}\}, \forall t'' \in \{1,2,\cdots,\nframes^{g_j}\}, \\
  &i,j=1,2, \cdots, \nvideos,
  \end{aligned}
  \label{problem_base}
\end{equation}
\emph{as defined in \eqref{problem_define}, 
$\dist$ denotes distance measurement such as commonly used Euclidean Distance}.

Different from the conventional ensemble learning approaches, 
where different base models are employed, 
here each base ranker for one video is embodied by the same re-identification 
network which receives different frames%
\footnote{Using different rankers, as done in conventional ensemble methods, 
is less convincing from our view because it significantly 
\cmm{increases the model complexity coming from multiple different models.}}.

By define the distance of the $t^{th}$ image $\samp_{t}^{p_i}$ of the $i^{th}$ 
query video $\data^{p_i}$ to the $j^{th}$ gallery set $\data^{g_j}$ as:
\begin{equation}
\begin{aligned}
  & d^{ij}_t=\min(d(\ranker(\samp_{t}^{{p_i}}),~\ranker(\samp_{t'}^{g_j})))), \\
  &\forall t' \in 1,2 \cdots \nframes^{g_j}~ and~ i,j=1,2 \cdots \nvideos,
\end{aligned}
\end{equation}

the base ranking list for the $i^{th}$ identity can be got by:

\begin{equation}
    \lists_{t}^i=\arg sort([d_t^{i1}, d_t^{i2},\cdots,d_t^{i\nvideos}]),
    \label{base_ranking}
\end{equation}

where $\arg sort$ is a ranking operator, 
which returns the indices of each video after ranking their distance with the 
$t^{th}$ image in the $i^{th}$ video ($\samp_{t}^{p_i}$). 
Without loss of generality, we assume that videos are ranked with 
increasing distance, i.e., the one with smallest distance gets the lowest rank. 
Each base ranking list in \eqref{base_ranking} is actually a permutation on 
$\{1,2,\cdots,\nvideos\}$. 
Moreover, each base ranking list $\lists_t^i$ is returned by different base 
ranker~$\ranker$ which is introduced in \secref{sec:baseranker}.

For each pair of video $(\data^{p_i},\data^{g_j})$ ($i,j=1,2 \cdots \nvideos$),
denoted by $\counts_{i,j}$ the times of $\lists_t^i$ that 
$i$ is ranked in front of $j$, that is  

\begin{equation}
   \counts_{i,j}= \sum_{t=1}^{\nframes^{p_i}} I(\lists^i_{t}(i) <  \lists^i_{t}(j)),
   \label{def:count}
\end{equation}

where $I$ is an indicator function ($I(x)=1$~i.f.f~$x==True$).  
We can further define that $\laby_{i,j}$ be true \emph{i.f.f.} 
$\counts_{i,j}>\frac{\nframes^{p_i}}{2}$. 
Moreover, suppose that the ground-truth ranking list for the $i^{th}$ video is 
$\hat{\lists^i}$, 
then based on Hoeffding’s inequality~\cite{serfling1974probability}, 
we have the following results:

\myPara{Proposition 1} 
\emph{Let 
$r=Pr((\lists_t^i(i)<  \lists_t^i(j))\neq (\hat{\lists^i}(i)<\hat{\lists^i}(j))$
denotes the error rate of each base ranker $ \lists_t^i$. 
Assume that $r<\frac{1}{2}-\varepsilon$,~$(0<\varepsilon<\frac{1}{2})$, 
under the i.i.d. assumption we have:\\}
\begin{equation}
  Pr(\laby_{i,j}=False|(\hat{\lists^i}(i)
  <\hat{\lists^i}(j))\leq e^{-2{\varepsilon}^2 \nframes^{p_i}}.
  \label{def:bound}
\end{equation}

Although the i.i.d. assumption for Hoeffding’s inequality may not hold in practice,  it allows for an analytical study of \task and sheds light upon how we could improve it. 
 

\myPara{Discussion on consistency} 
For the $i^{th}$ gallery video $\data^{g_i}$, 
in order to aggregate multiple results from each base ranker 
$\ranker^i_{t}, t=1,~2,\cdots,\nframes^{g_i}$, 
variables $\laby_{i,~j}$ must define a transitive relation, 
that is $\laby_{i,~k}$ is \emph{True} if both $\laby_{i,~j}$ and $\laby_{j,~k}$ 
are \emph{True}. 
When this is held for all $i,~j,~k$, base rankers 
$\ranker^i_{t}, t=1,2,\cdots,\nframes^{g_i}$ are said \emph{consistent}
\cite{jong2004ensemble}. 
Theoretical analysis in~\cite{jong2004ensemble} shows that under the 
same condition (i.i.d and each base ranker is better than a random guess),  
$\ranker^i_{t},t=1,~2,\cdots,\nframes^{g_i}$  are consistent with 
probability 1 as $\nframes^{g_i}$ goes to infinity.
This naturally guarantees us to \emph{Kemeny–Young method} 
\cite{levin1995introduction} to aggregate the final ranking list. 
Specifically, \emph{Kemeny–Young method} uses preferential ballots 
on each base ranker $\ranker^i_{t}, t=1,~2,\cdots,\nframes^{g_i}$. 
We first create a 0-1 binary matrix 
$\Counts \in \{0,~1\}^{{\nvideos}\times{\nvideos}\times{\nframes^{p_i}}}$, 
where $\Counts_{ijt}=1$ ($i,~j=1,~2 \cdots \nvideos$ and 
$t=1,~2 \cdots \nframes^{g_i})$ \emph{i.f.f} $R^i_{t}(i)<R^i_{t}(j)$. 
Otherwise, $\Counts_{ijt}=0$. Then a summation along the $2^{nd}$ 
and $3^{th}$ dimensions returns a vector $\Counts_{i\cdot \cdot}$, 
where the $k^{th}$ entry $\Counts_{i\cdot \cdot }(k)$ summarizes the frequency 
that the $k^{th}$ identify in the gallery set is favoured in all the base rankers.
The final ranking list is obtained by ranking the vector $\Counts_{i\cdot \cdot }$ 
in a decreasing manner.

\subsection{Base Ranker}
\label{sec:baseranker}


Generally, our solution is widely applicable to any image based person 
re-identification methods. 
However, as indicated in \eqref{def:bound} from Proposition $2$,  
under the i.i.d. assumption, 
the error bound of the final ensemble decreases exponentially with the square 
of $\varepsilon$. 
Hence it would be beneficial in practice to use the best-performing base ranker.
Motivated by this, we embark on the state-of-the-art image based person 
re-identification approaches. 
To illustrate our point, we present two realizations of our proposed method:
\emph{EnsembleRanker-RW} and \emph{EnsembleRanker-CRF}. 
The former uses random walk based methods in \cite{shen2018deep} as base ranker 
while the later employs CRF based methods in \cite{chen2018group}. 
We choose these two methods because of their superior performances for 
image based person re-identification. 
Below their main techniques are briefly introduced and interested readers 
are refereed to the original paper for more details.

\myPara{Ensemble-RW} 
This method trains the person re-identification model with the technique in
\cite{shen2018deep}. 
More specifically, it adopts a novel group-shuffling random walk operation 
for fully utilizing the affinities between gallery images to refine 
the affinities between probe and gallery images. 
It integrates random walk operation into the training process of 
deep neural networks. 
Apart from the ground-truth identify label, 
richer supervision could be provided by grouping and shuffling the features 
through the random walk operation. 
It achieves a top-1 accuracy of $91.5$, $92.7$, and $80.7$ on CUHK03, 
Market-1501, and DukeMTMC, respectively~\cite{shen2018deep}.

\myPara{Ensemble-CRF} 
This method employs the technique in \cite{chen2018group}. 
It uses a novel similarity learning approach for person re-identification 
by combining the CRF model with deep neural networks. 
It models the similarities between images in the group via a unified graphical model,
and learns local similarities with the aid of group similarities in 
a multi-scale manner. 
As more inter-image relations are considered, 
the learned similarity metric is reported to be more robust and consistent 
with images under challenging scenarios. 
This method achieves an top-1 accuracy of $90.2$, $93.5$, and $84.9$ on CUHK03,
Market-1501, and DukeMTMC, respectively \cite{chen2018group}.

\subsection{Difference with conventional ensemble learning}

Existing ensemble re-id methods~\cite{prosser2010person,paisitkriangkrai2015learning}
mainly come from image based task point of view.  
The work in \cite{prosser2010person} trains the ensemble RankSVM, 
while methods in \cite{paisitkriangkrai2015learning} uses multiple features 
to get multiple metrics and combine them through weighted average. 
In contrast, we investigate the feasibility of ensemble learning for \task 
under the umbrella of deep learning in which the efficiency and the robustness 
is guaranteed by a shared base ranker and  the KemenyYoung method, respectively. 
More importantly, we show the effectiveness of the proposed method both 
empirically and theoretically. 
Other conventional ensemble methods~\cite{ren2016ensemble},  
which typically employ different base models \cmm{and yield much larger model complexity}, 
and hence are less interesting to investigate here.

\section{Experimental Results}

In this section, we evaluate our proposed approach to video re-identification
on three challenging benchmark datasets: iLIDS-VID~\cite{wang2014person},
PRID-2011~\cite{hirzer2011person}, and MARS~\cite{zheng2016mars}. 
The iLIDS-VID dataset contains $300$ persons, 
where each person is represented by two video sequences captured by 
non-overlapping cameras. 
The PRID-2011 dataset contains $749$ persons, 
captured by two non-overlapping cameras. 
Following the protocol used in~\cite{wang2014person}, 
sequences with more than $21$ frames are selected, leading to $178$ identities. 
For iLIDS-VID and PRID-2011, each dataset is randomly split into 
$50\%$ of persons for training and $50\%$ of persons for testing. 
The MARS dataset is the largest video-based person re-identification benchmark 
with $1,261$ identities and around $20,000$ video sequences generated by 
DPM detector \cite{felzenszwalb2010object} and GMMCP tracker~\cite{dehghan2015gmmcp}.
Each identity is captured by at least $2$ cameras and has $13.2$ sequences 
on average. 
There are $3,248$ distractor sequences in the dataset.

For Ensemble-RW, the network is trained for $40$ epochs using stochastic 
gradient descent with a learning rate of $1e-6$, and a batch size of $32$. 
For Ensemble-CRF, the network is trained for $100$ epochs using stochastic 
gradient descent with a learning rate of $1e-2$, and a batch size of $132$. 
The other hyper-parameters are set to the same values as the original one. 
We train the network with two Nvidia TiTan GPUs. 
For the first two datasets, during testing we consider the first camera 
as the probe and the second camera as the gallery \cite{wang2014person}. 
For MARS, we follow exactly the same evaluation protocol in \cite{zheng2016mars}. 
A probe sequence is passed through the network to get the feature vector and 
ranked using Euclidean distance with pre-computed feature vectors from all 
gallery sequences. 
Then we use the \emph{KemenyYoung} method to aggregate the results for 
a final ranking list.

\renewcommand{\addImg}[2]{\subfloat[#1]{\includegraphics[height=0.265\linewidth]{#2}}}

\begin{figure}
  \centering 
  \addImg{PRID-2011}{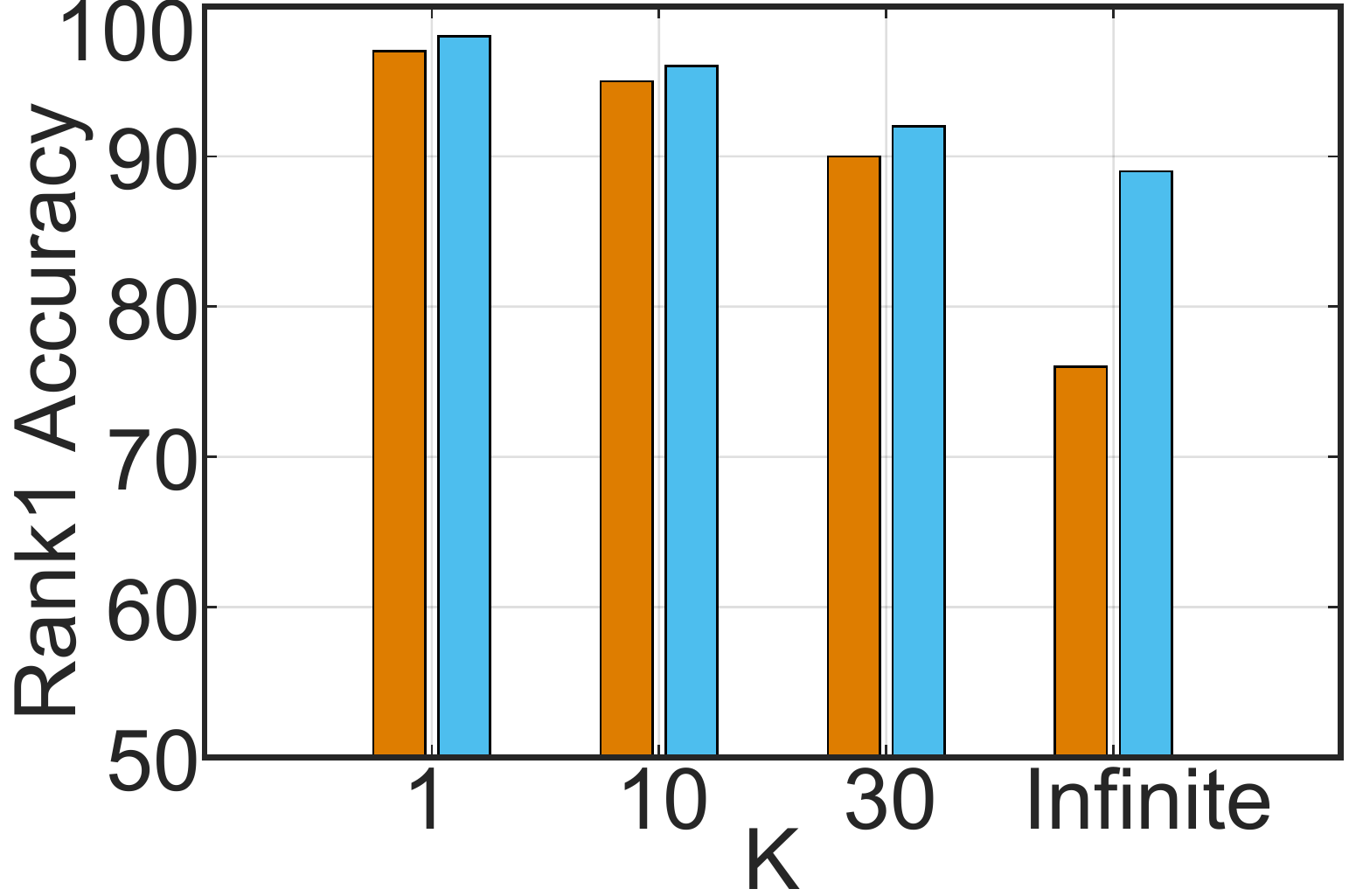} \hfill
  \addImg{iLIDS-VID}{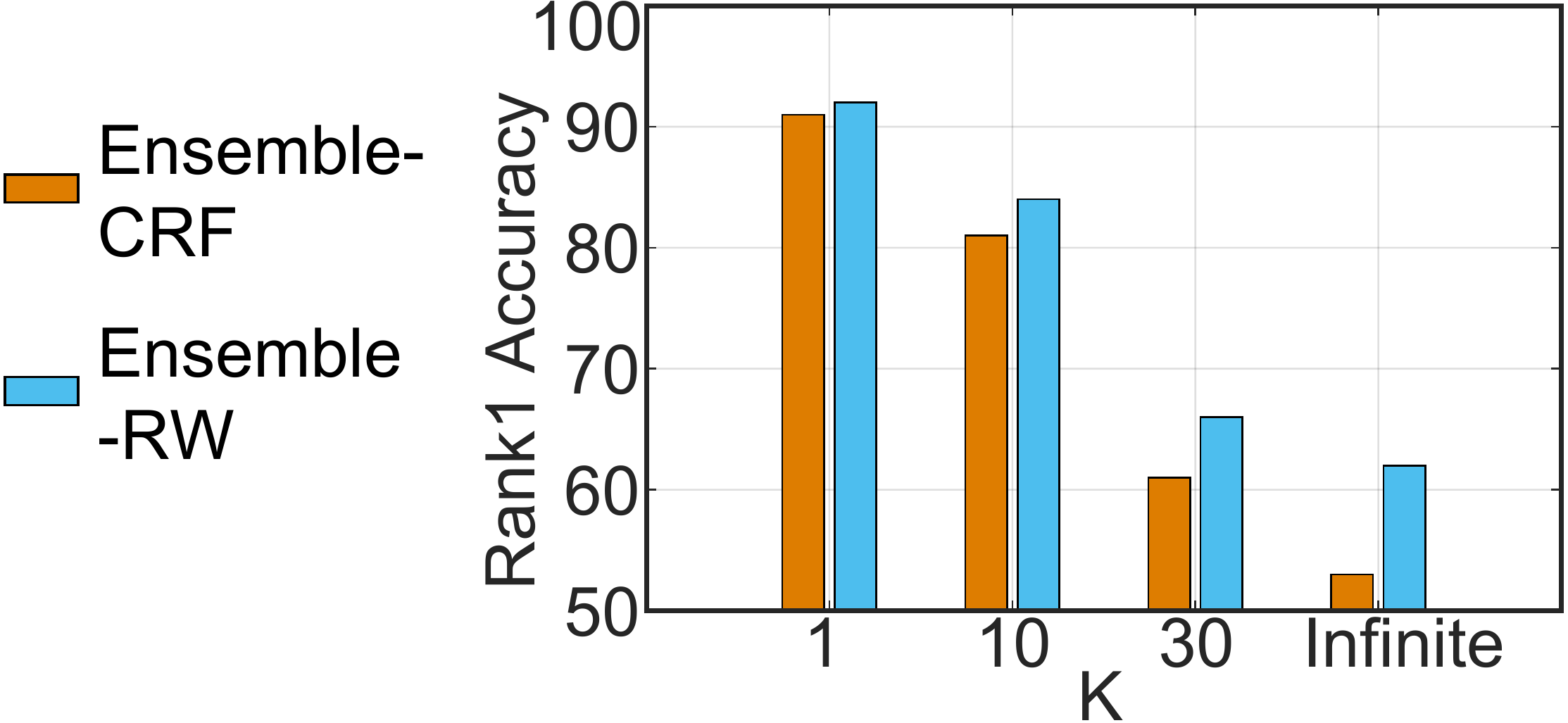}  \\
  \caption{\textbf{CMC Rank 1 accuracy of Ensemble-RW and Ensemble-CRF 
    for different sample rate $K$.} 
    We show how the sample rate impacts the accuracy of our solution. 
    We report the results of $K=1, 10 , 30$, and $Infinite$, 
    in which $K=Infinite$ stands for the case where only the first image 
    in the video is sampled. 
    The experimental results show that the accuracy increases 
    with the decrements of $K$.
  }\label{fig-frame}    
\end{figure}

\subsection{Ablation Studies}

We first provide some ablation studies to investigate two key aspects in
\eqref{def:bound}. 
We conduct those experiments on iLIDS-VID and PRID-2011.

\myPara{Effect of $\nframes$:}
Our theoretical analysis shows that the error bound of the final system decreases
exponentially with the number of image frames under the certain conditions. 
However, in practice we cannot leverage unlimited length of video sequences 
due to constrained storage, transmission as well as computation resources. 
In order to investigate the effect of the video length, 
we report the results both Ensemble-EW and Ensemble-CRF on PRID-2011 
and iLIDS-VID dataset. 
More specifically, we sample one image from every $K$ images in each video sequences. 
We set $K=Infinite$ to make sure  $K\geq \nframes$, and in this case, 
we use the first frame to conduct the image based person re-identification 
experiment as done in~\cite{zheng2016mars}. 
When $K=1$, we use all the available frames. Results are illustrated in 
\figref{fig-frame}. 
It is obvious that $K$ has an inverse effect on the final performance, 
that is, the accuracy increases with the decrements of $K$.

With respect to the ``probe-to-gallery" pattern, person re-identification can 
be mainly categorized into two strategies: image-to-image and video-to-video. 
The first mode is mostly studied in literature while only recently, the video-to-video re-id has been investigated.  
Other scenarios such as image-to-video and video-to-image are less common 
and can be regarded as a special case of video-to-video.  
Our findings apparently verify the the common belief that  
the video-to-video pattern, which is also the focus in this paper, 
is more favourable because both probe and gallery units contain much richer 
visual information than single images.  
In the following section further show that other common practices 
such as modelling temporal dynamics in videos for~\task may not help 
to improve the performance.

\begin{figure}
  \centering 
  \addImg{PRID-2011}{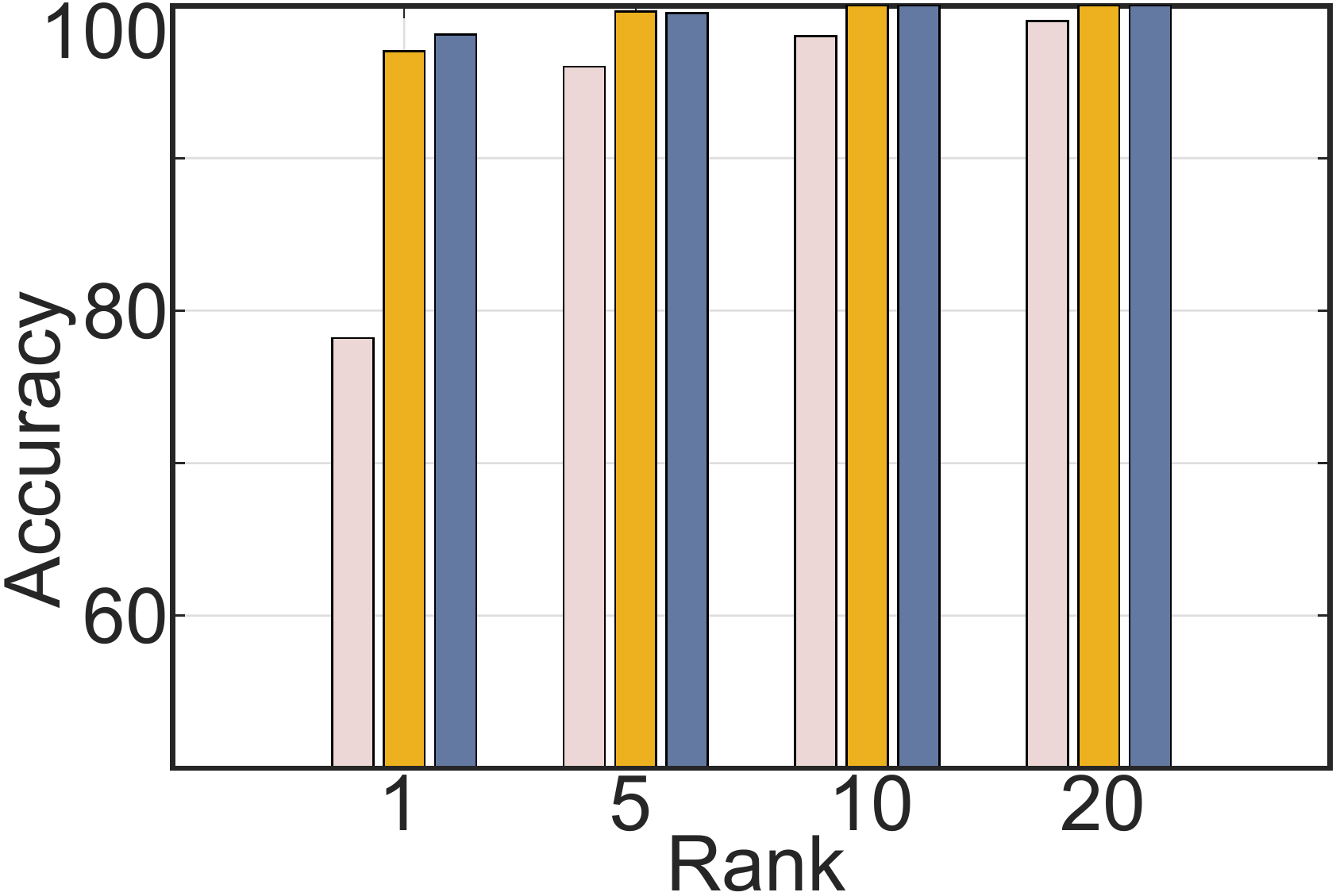} \hfill
  \addImg{iLIDS-VID}{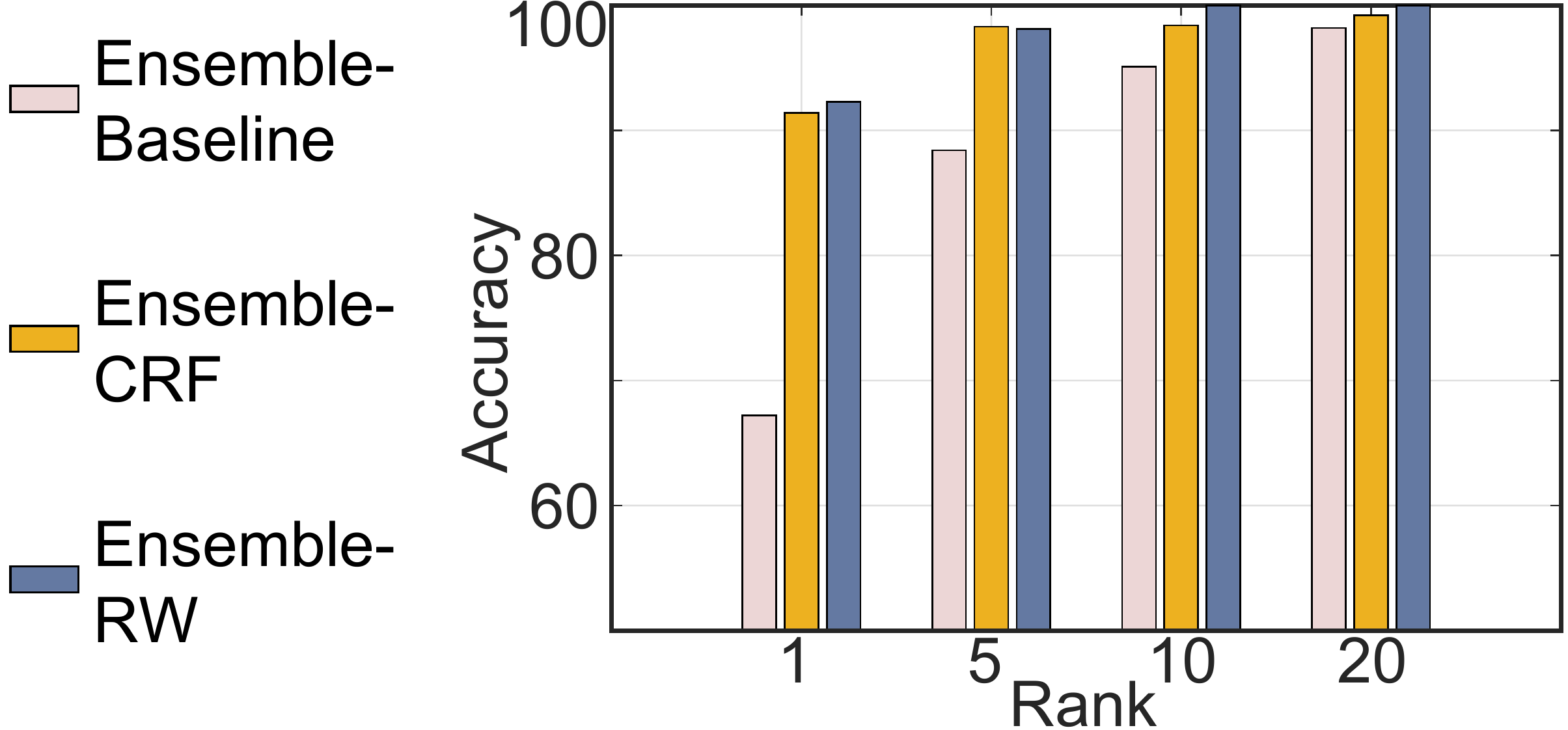}  \\
  \caption{\textbf{Comparison of different base rankers.} 
    We report the results of three different rankers. 
    The baseline method in \cite{shen2018deep}, 
    which disable the functionality to refine the affinities between probe 
    and gallery images, 
    is integrated in our system as ``Ensemble-Basline". 
    Our results on two datasets show that an ensemble of  
    stronger base ranker perform better.
  }\label{fig-ranker}    
\end{figure}

\myPara{Effect of $\varepsilon$:} 
The error bound in \eqref{def:bound} shows improving the final accuracy from 
another point of view, that is, reducing the error of each base ranker. 
To show this, we train the baseline model done in~\cite{shen2018deep}.  
This method does not use group-shuffling random walk operation and thus 
the ability of refining the affinities between probe and gallery images is disabled. 
This version has been reported to be worse in all the dataset used in 
\cite{shen2018deep}. 
We call this method ``Ensemble-Baseline" and the results in  
are summarized in \figref{fig-ranker}. 
It is obvious that using stronger base rankers can lead to better overall 
accuracy in all cases.

\begin{figure}[t]
  \scriptsize
  \centering
  \setlength{\tabcolsep}{2.1mm}
  \begin{tabular}{@{}lcccccc@{}} \toprule
    & Rank 1 & Rank 5 & Rank 20 & MAP \\ \hline
    Liu \etal~\cite{liu2018video}& 68.3&81.4&90.6&52.9\\
    Zheng \etal~\cite{zheng2016mars}& 68.3&82.6&89.4&49.3\\
    Zhou \etal~\cite{zhousee}& 70.6&90.0&97.6&50.7\\
    Chen \etal~\cite{chen2017deep}& 71.0&89.0&96.0&-\\
    Zhang \etal~\cite{zhang2018multi}&71.2&85.7&94.3&-\\
    Li \etal~\cite{li2018diversity}& 82.3&-&-&65.8\\
    Wu \etal~\cite{wu2018exploit}&80.8&92.1&96.1&67.4\\
    \zl{Zhang \etal~\cite{zhang2018multi}}&71.2&85.7&94.3&-\\
    \hline
    Ensemble-CRF& {85.3}&\textbf{95.0}&{97.1}&{79.3}\\
    Ensemble-RW& \textbf{86.8}&{94.6}&\textbf{97.5}&\textbf{80.1}\\
    \hline 
    \zl{Ensemble-CRF (MC)}&85.1&94.8&97.0&79.2\\
    \zl{Ensemble-RW (MC)}&86.3&94.4&97.4&\textbf{80.1}\\
    \bottomrule
  \end{tabular}
  \caption{\textbf{Comparison  with other state-of-the-art methods on MARS.}}
  \label{tab_mars}
\end{figure}

\subsection{Comparisons with State-of-the-Art}

\begin{figure*}[!thb]
  \centering
  \scriptsize
  \setlength{\tabcolsep}{4.0mm}
  \begin{tabular}{@{}lcccccccc@{}} \toprule
    &\colsN{4}{PRID-2011} & \colsN{4}{iLIDS-VID} \\ 
    \cmidrule(lr){2-5} \cmidrule(lr){6-9}
    & Rank 1 & Rank 5 & Rank 10 & Rank 20 & Rank 1 & Rank 5 & Rank 10 & Rank 20\\\hline
    Karanam \etal~\cite{karanam2015sparse}
    &35.1&59.4&69.8&79.7&24.9&44.5&55.6&66.2\\
    Karanam \etal~\cite{karanam2015person}
    & 40.6&69.7&77.8&85.6&25.9&48.2&57.3&68.9\\
    Wang \etal~\cite{wang2014person}
    &41.7&64.5&77.5&88.8&34.5&56.7&67.5&77.5\\
    Li \etal~\cite{li2015multi}
    &43.0&72.7&84.6&91.9&37.5&62.7&73.0&81.8\\
    Wu \etal~\cite{wu2016deep}
    &49.8&77.4& 90.7& 94.6&42.6& 70.2&86.4& 92.3\\
    Li \etal~\cite{li2017video}
    &51.6& 83.1& 91.0 &95.5&34.5& 63.3& 74.5&84.4\\
    Yan \etal~\cite{yan2016person}
    &58.2 &85.8& 93.4&97.9&49.3& 76.8&85.3& 90.0\\
    Liu \etal~\cite{liu2015spatio} 
    &64.1&87.3&89.9&92.0&44.3&71.7&83.7&91.7\\
    McLaughlin \etal~\cite{mclaughlin2016recurrent}
    &70.0 &90.0 &95.0 &97.0& 58.0 &84.0 &91.0& 96.0\\
    Zhang \etal~\cite{zhang2017learning}
    &72.8& 92.0& 95.1& 97.6&55.3& 85.0& 91.7& 95.1\\
    Chen \etal~\cite{chen2017deep}
    &77.0& 93.0& 95.0& 98.0& 61.0&85.0& 94.0& 97.0\\
    Xu \etal~\cite{xu2017jointly}
    &77.0& 95.0& 99.0& 99.0&62.0& 86.0& 94.0& 98.0\\
    Zhou \etal~\cite{zhousee}
    &79.4&94.4&-&99.3&55.2 &86.5&-& 97.0\\
    Liu \etal~\cite{liu2018video}
    &83.7&98.3&99.4&\textbf{100.0}&68.7&94.3&98.3&99.3\\
    Zhang \etal~\cite{zhang2018multi}
    &85.2&97.1&98.9&99.6&60.2&84.7&91.7&95.2\\
    Liu \etal~\cite{liu2017quality}
    &90.3&98.2&99.3&100.0&68.0&86.8&95.4&97.4\\
    Khan \etal~\cite{khan2017multi}
    &92.5&99.3&\textbf{100.0}&\textbf{100.0}&79.5&95.1&97.6&99.1\\
    Li \etal~\cite{li2018diversity}
    &93.2&-&-&-&80.2&-&-&-\\
    \zl{Zhang \etal~\cite{zhang2018multi}}
    &85.2&97.1&98.9&99.6&60.2&84.7&91.7&95.2\\
    \hline
    Ensemble-CRF&{95.5} &\textbf{99.6}&\textbf{100.0}
    &\textbf{100.0}&{90.4}&\textbf{98.3}&{98.4}&{99.2}\\
    Ensemble-RW&\textbf{97.1}&{99.5}&\textbf{100.0}&\textbf{100.0}
    &\textbf{91.3}& {98.1}&\textbf{100.0}&\textbf{100.0}\\
    \hline
    \zl{Ensemble-CRF (MC)}&94.7&98.9&\textbf{100.0}&\textbf{100.0}&89.7&97.7&97.6&99.1\\
    \zl{Ensemble-RW (MC)}&96.8&99.3&\textbf{100.0}&\textbf{100.0}&\textbf{91.3}&97.9&\textbf{100.0}&\textbf{100.0}\\
    \bottomrule        
  \end{tabular}
  \caption{\textbf{Comparison with other state-of-the-art methods on PRID-2011 
    and iLIDS-VID dataset.
  }}\label{tab:sota-compare}
\end{figure*}

Notable recent works on~\task, 
that have improved the state-of-the-art results are: 
1) McLaughlin \etal's~\cite{mclaughlin2016recurrent} method that 
uses temporal average pooling to aggregate RNN outputs at each time step; and 
2) Chen \etal's method~\cite{chen2017deep} that
adopts the similar network structure with~\cite{mclaughlin2016recurrent}, 
but fuses both CNN and RNN features; and 
3) Xu \etal's~\cite{xu2017jointly} that uses spatial pyramid pooling in CNN 
and attention models for more discriminative features; 
4) Zhang \etal's~\cite{zhang2017learning} that integrates CNNS and BRNNs; 
5) Zhou \etal's method~\cite{zhousee} that uses both temporal attention model,
recurrent units and more complicated triplet loss; 
6) Liu \etal's method which uses AMOC network jointly learns appearance 
representation and motion context from a collection of adjacent frames 
using a two-stream convolutional architecture. 
7) Li \etal's method~\cite{li2018diversity} that uses spatial attention ensemble 
to discover the same body part. 
8) Zhang \etal's method~\cite{wu2018exploit} that employs an interpretable
reinforcement learning to train an agent to verify a pair of images at each time.
From \tabref{tab:sota-compare}, we can observe that both the proposed methods
outperform ``CNN-RNN" and its several advanced extensions.  \tabref{tab_mars} compares the proposed method with other state-of-the-arts 
methods on MARS. 
Following the standard evaluation protocol, 
we resort to both CMC and MAP as the evaluation metric~\cite{zheng2016mars}. 
The Average Precision (AP) is calculated based on the ranking result, 
and the mean Average Precision (mAP) is computed across all queries 
which is viewed as the final re-identification accuracy. 
For the ease of presentation, we omit the results of CMC Rank 15. 
Again, our proposed solution outperforms all those methods in terms of CMC Rank 
and MAP scores. 
\zl{In addition, we also include other variants of the proposed method, 
which use the Cross-Entropy Monte Carlo algorithm~\cite{pihur2009rankaggreg} 
to aggregate multiple rankers.  
We observe that with other aggregation methods, our proposed solutions still
outperform all those methods in terms of CMC Rank and MAP scores.}
 
 Here we provide some open discussions as well.~\eqref{def:bound} suggests a good way to improve person re-identification
performance: 
using the best image person re-identification methods on all the 
available video frames. 
However, this is non-trivial and deserves future research work 
in terms of 
%
the following problems: 
1) How to satisfy the i.i.d.\  assumption? and 
2) given a limited computational budget, how to achieve a good trade-off between 
$\nframes$ and $\varepsilon$? 
These may be investigated in various aspects: 
(1) Enhanced dataset precessing in which dependencies among
individual frames can be reduced, such as random frame skip or key frame selection 
\cite{meng2016keyframes}. 
However, this also reduces the video length. 
(2) Using more compute-intensive methods is beneficial in improving $\varepsilon$. 

\section{Conclusion}
We first provide a diagnostic analysis on the commonly used ``CNN-RNN" recipes 
for \task. 
We show that RNNs used in the literature usually \zl{may not be effective} to capture the temporal
dependencies and implicitly learn an orderless representation, 
which we believe is more pertinent for \task.
Based on this observation, we then propose a simple yet surprisingly powerful ensemble
approach for \task. 
Moreover, we theoretically prove that both employing strong image based person 
re-identification methods and long video sequences are beneficial. 
We demonstrate this idea with two examples and our proposed solution 
significantly outperforms the existing state-of-the-arts methods, 
on multiple widely used datasets.

\bibliographystyle{IEEEtran}
\bibliography{egbib}

\newcommand{\AddPhoto}[1]{\includegraphics%
[width=1in,keepaspectratio]{#1}}

\vfill

\end{document}